\documentclass{article}
\usepackage{spconf,amsmath,graphicx}

\usepackage{setspace}
\usepackage{algorithm}
\usepackage{algorithmic}
\title{Geometric Data Augmentation Based on Feature Map Ensemble}
\name{Takashi Shibata$^{1}$, Masayuki Tanaka$^{2}$, Masatoshi Okutomi$^{2}$}
\address{$^{1}$NTT Corporation, $^{2}$Tokyo Institute of Technology}

\begin{document}
%
\maketitle
\begin{abstract}
Deep convolutional networks have become the mainstream in
computer vision applications. Although CNNs have been successful in many computer vision tasks, it is not free from drawbacks. The performance of CNN is dramatically degraded by geometric transformation, such as large rotations. In this paper, we propose a novel CNN architecture that can improve the robustness against geometric transformations without modifying the existing backbones of their CNNs. The key is to enclose the existing backbone with a geometric transformation (and the corresponding reverse transformation) and a feature map ensemble. The proposed method can inherit the strengths of existing CNNs that have been presented so far. Furthermore, the proposed method can be employed in combination with state-of-the-art data augmentation algorithms to improve their performance. We demonstrate the effectiveness of the proposed method using standard datasets such as CIFAR, CUB-200, and Mnist-rot-12k.

\renewcommand{\thefootnote}{\fnsymbol{footnote}}
\footnote[0]{©2021 IEEE. Personal use of this material is permitted. Permission from IEEE must be obtained for all other uses, in any current or future media, including reprinting/republishing this material for advertising or promotional purposes, creating new collective works, for resale or redistribution to servers or lists, or reuse of any copyrighted component of this work in other works. }
\renewcommand{\thefootnote}{\arabic{footnote}}

\end{abstract}
\begin{spacing}{0.965}

\section{Introduction}
\label{sec:intro}
The success of image recognition benchmarks on ImageNet~\cite{russakovsky2015imagenet} has triggered the significant progress of convolutional neural networks (CNNs) ~\cite{he2016deep,tan2019efficientnet,zagoruyko2016wide,hu2018squeeze,huang2017densely}. 
Those CNN architectures and their variants are widely applied to various computer vision tasks~\cite{long2015fully,ren2015faster}. 
Even though CNNs have established as a powerful framework, they are not entirely free from drawbacks. 
It is widely known that the performance of CNNs is dramatically degraded by geometrical transformation, such as large-scale rotations. 

The typical way to handle this degradation is to employ data augmentation. 
Recently, more sophisticated data augmentation algorithms based on reinforcement learning and random search have been proposed~\cite{cubuk2020randaugment,cubuk2018autoaugment,lim2019fast}.
Even though these augmentation algorithms can obtain a highly expressive model, training becomes significantly difficult due to a large number of trials when their magnitude of data augmentation (e.g., rotation angle range) is increased.

	\begin{figure*}[t]
        \centering\includegraphics[width=0.6\linewidth]{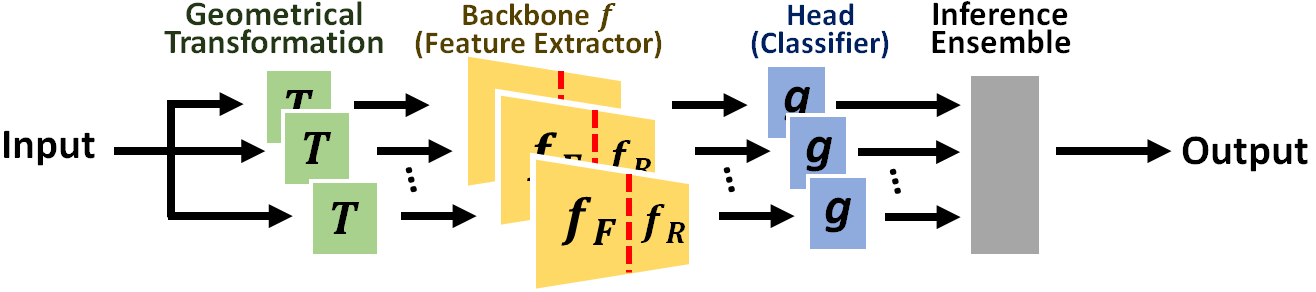} \\
        {\vspace{-0.1cm} \small{(a) Standard CNN with TTA \vspace{0.2cm}}}\\
        \centering\includegraphics[width=0.88\linewidth]{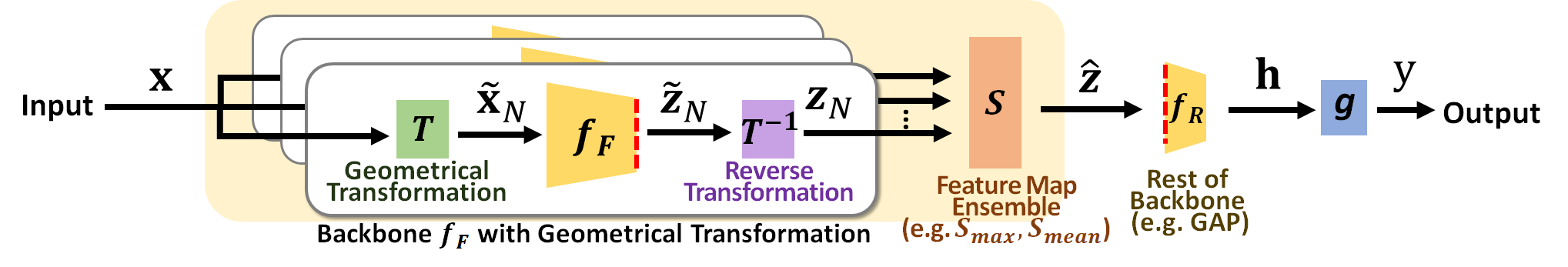} \\
        {\vspace{-0.1cm} \small{(b) Proposed method 
        with forward and reverse transformations
        \vspace{0.2cm}}}\\
        \vspace{-3mm}
		\caption{{\bf{Overview of the proposed method.}} The proposed method improves the robustness to geometric variations such as rotation by ensembling the set of the feature maps from the backbone $f_F$.}
		\label{fig:proposed pileline2}
	\end{figure*}

The data augmentation is also useful at test time, called Test-Time Augmentation (TTA). Test-Time Augmentation makes the recognition accuracy more robust by computing the mean or maximum inference scores. 
However, the performance tends to drop dramatically in the case of false acceptance with high inference score because the final recognition accuracy is only evaluated based on the inference scores. 
Furthermore, it has also been pointed out that TTA can turn many correct predictions into wrong predictions, even if it produces a model with improved overall accuracy~\cite{shanmugam2020and}.
It suggests that the performance gain is limited when the input and the inference scores are only manipulated. In other words, it is necessary to incorporate the data augmentation framework into the network structure.

In this paper, we propose a novel CNN architecture that can improve the robustness against geometric transformations without modifying the existing backbones of the original CNN. 
The key is to enclose the existing backbone with a geometrical transformation (and the corresponding reverse transformation) and a feature map ensemble.
As shown in Fig.~\ref{fig:proposed pileline2}~(a), in a typical TTA, the final output is obtained by integrating the inference scores. 
On the other hand, the proposed method integrates the feature maps at the intermediate layer, as shown in Fig.~\ref{fig:proposed pileline2}~(b).
The proposed method is inspired by the Maxout networks~\cite{goodfellow2013maxout}, which was proposed to train complex activation functions.
Other existing methods~\cite{laptev2016ti,zhang2018mintin} also use the Maxout architecture in a fully connected layer to obtain rotation-invariant features.
In contrast, the proposed method takes advantage of the Maxout architecture for integrating the feature maps corresponding to each data augmentation. 
The implementation of our method is very simple and can be implemented with only a few changes to the conventional CNN architecture. 
 Moreover, our method can be used in combination with state-of-the-art data augmentation techniques. 
 We demonstrate the effectiveness of the proposed method using standard datasets, including CIFAR, CUB-200, and Mnist-rot-12k.

\section{Proposed method}
\label{sec:proposed method}
\subsection{Processing pipeline of the proposed framework}
Our goal is to improve the robustness to geometrical transformation without modifying the original backbone of the CNN architecture
so that we can reuse the high-performance pretrained backbone.
We begin our discussion with a standard CNN with TTA shown in Fig.~\ref{fig:proposed pileline2}~(a).
The standard CNN contains the backbone and the head.
Let ${\bf{x}}$, $f(\cdot)$, and $g(\cdot)$ be an input image, the backbone, and the head, respectively. 
The backbone ${f}$ further consists of ${f_{F}}$ and ${f_{R}}$. 
The output of the backbone ${f_{F}}$ 
generates the feature map ${\bf{z}}=z_{chw}$, where $c$, $h$, and $w$ are the indexes of  channel, height, and width, respectively.
The output $y$ for input image ${\bf{x}}$ is given by $y=g(f({\bf{x}}))=g(f_{R}({\bf{z}}))=g(f_{R}(f_{F}({\bf{x}})))$. 
In the typical TTA manner, the inference scores are combined by evaluating the mean or maximum.

On the other hand, as shown in Fig.~\ref{fig:proposed pileline2}~(b), the proposed method consists of 1) the backbone with 
forward and reverse geometric transformations, 2) the feature ensemble, and 3) the head. In the following, we describe 
each component.

\vspace{-0.3cm}
\subsubsection{Backbone with geometrical transformations}
The proposed method generates the set of the aligned feature maps by enclosing the original backbone $f_F$ with a geometric transformation $T$ and the corresponding reverse transformation $T^{-1}$.

\vspace{0.1cm}
\noindent
{\bf{Geometrical transformation}}: 
First, we generate a set of the geometrical transformed images 
$ \mathcal{\tilde{X}} = \{ {\tilde{{\bf{x}}}}_{1} , \cdots, {\tilde{{\bf{x}}}}_{n},\cdots, {\tilde{{\bf{x}}}}_{N} \}$ 
from the input $ {\bf{x}} $, where $n$ is the index for each geometrical transformation, $N$ is the number of the geometrical transformation. The $n$-th transformed image ${\tilde{{\bf{x}}}}_{n}$ is given by
\begin{equation}
{\tilde{{\bf{x}}}}_{n} = {\it{T}}( {\bf{x}};  {\xi}_n   ),
\end{equation}
\noindent
where $ {\xi}_n $ is a parameters for the $n$-th geometrical transformation. 
In this paper, four rotations with discrete angles (i.e., $0$~[deg], $90$~[deg], $180$~[deg] and $270$~[deg]) are used as the geometrical transformation represented as $\{ {\xi}_n \} = \{ {\bf{R}}_{0}, {\bf{R}}_{90}, {\bf{R}}_{180}, {\bf{R}}_{270} \}$, where  ${\bf{R}}_{\theta}$ is the rotation matrix that rotates the input $ {\bf{x}}$ by ${\theta}$ degrees. 
Note that the proposed method can deal with various geometrical transformations  instead of the discrete rotations. 
The set of the transformed images $ \mathcal{\tilde{X}} $ is used as the input of $f_F$, and then the set of the feature maps with the geometrical transformation is obtained as  
$ \mathcal{\tilde{Z}} = \{ {\tilde{{\bf{z}}}}_{1}, {\tilde{{\bf{z}}}}_{2} \cdots, {\tilde{{\bf{z}}}}_{N} \} = \{ f_F({\tilde{{\bf{x}}}}_{1}), f_F({\tilde{{\bf{x}}}}_{2}) \cdots, f_F({\tilde{{\bf{x}}}}_{N}) \}$.

\vspace{0.1cm}
\noindent
{\bf{Reverse transformation}}: 
The reverse transformation geometrically reverses the set of the feature maps $\mathcal{\tilde{Z}}$.
As described later, the feature map 
ensemble selects one feature from the set of features corresponding to each position. 
To compare the feature maps with the same positions, these feature maps should be 
geometrically 
aligned by the reverse transformation. 
The parameters for this reverse transformation are uniquely determined from the geometrical transformation parameters $\{ {\xi}_n \}$. 
The set of the reverse transformed feature maps is denoted as  $ \mathcal{{Z}} = \{ {\bf{z}}_{1},\cdots, {\bf{z}}_{n} \cdots, {\bf{z}}_{N} \}$, where ${\bf{z}}_{n}$ is the $n$-th reverse transformed feature map  from the input ${\bf{x}}$, given by   
\vspace{-0.3cm}
\begin{eqnarray}
\label{eq:backbone}
{\bf{z}}_{n} 
= {\it{T}}^{-1}( f_F(  {\it{T}}( {\bf{x}};  {\xi}_n  )  ) ;  {\xi}_n ). 
\end{eqnarray}

\vspace{-0.6cm}
\subsubsection{Feature map ensemble}
\vspace{-0.1cm}
Feature map ensemble integrates the set of  the reverse transformed feature maps $\mathcal{{Z}}$. 
To integrate the feature maps, we concatenate the set of feature maps $\mathcal{{Z}}$, and obtain the concatenated feature maps ${\bf{Z}}=[ {\bf{z}}_{1},\cdots, {\bf{z}}_{n} \cdots, {\bf{z}}_{N} ]$, where $[ \cdot  ]$ is concatenation operator. In indax notation, the concatenated feature maps ${\bf{Z}}$ is represented as $Z_{nchw}$. 
In this paper, two types of ensemble algorithms, called the feature map maximum and the feature map mean, are proposed as follows: 
\begin{eqnarray}
&{\hat{\bf{z}}}&=S_{{\rm{max}}}({\bf{Z}}) = \max_n Z_{nchw} , \\ 
&{\hat{\bf{z}}}&=S_{{\rm{mean}}}({\bf{Z}})= \frac{1}{N} \sum_{n} {\bf{z}}_{n}.
\end{eqnarray}
The final output inference $y$ is given by ${y} = g(f_{R}({\hat{\bf{z}}}))$.

Note that the existing methods~\cite{laptev2016ti,zhang2018mintin} 
use the Maxout architecture at the fully connected (fc) layers. Therefore, these methods are only applicable to image classification task because the spatial information is already lost at the fc layers. In contrast, the proposed method has the potential to apply various tasks such as semantic segmentation because the proposed method integrates the feature maps containing the spatial information.

\vspace{-0.15cm}
\subsection{Analysis}
\label{sec:ana}
\vspace{-0.15cm}
We demonstrate the advantages of the proposed feature map ensemble using toy examples. 
The rotated training images were generated by randomly rotating the original CIFAR10 to one of four angles  (0~[deg], 90~[deg], 180~[deg], and 270~[deg]). 
On the other hand, the four sets of rotated test images were generated by rotating the original test image of CIFAR10  with all four angles (i.e., 0~[deg], 90~[deg], 180~[deg], and 270~[deg]). 
 We used ResNet18~\cite{he2016deep}  for this analysis. 
 The weight of ResNet18 was obtained by fine-tuning from the pre-trained model whose weight is optimized using the original CIFAR 10 without rotation. 

\begin{figure}[t]
        \centering\includegraphics[width=0.99\linewidth]{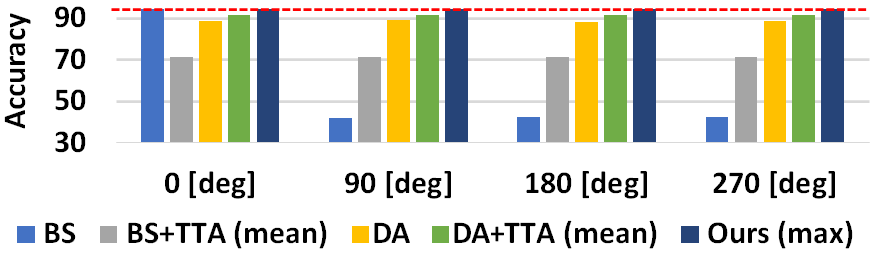}
        \vspace{-0.2cm}
		\caption{{\bf{Results on CIFAR10 with  four discrete rotations.}}}
		\label{fig:A}
\end{figure}

Figure~~\ref{fig:A} shows the comparison of our algorithm to four algorithms, where BS represents a base model trained with original images only, DA represents a model trained with images augmented by rotating, and TTA represents four angle TTA.
We used max operation for our algorithm and mean operation for TTA to obtain a better performance.
The accuracy of BS
is significantly decreased by the angle rotation. 
Even if the data augmentation and the TTA are performed, 
the accuracy is lower than that of BS with the original CIFAR10 (i.e., the BS's accuracy with 0 [deg] shown as the red dotted line).
On the other hand, our proposed method can keep accuracy among all rotations. These results demonstrate the robustness of the proposed method to the rotation.

\vspace{-0.15cm}
\section{Experiments}
\label{sec:intro}
\vspace{-0.15cm}
We first compared the performance of the proposed method with that of test-time augmentation (TTA).
Then, we demonstrate the effectiveness of the proposed method using Mnist-rot-12k dataset. Finally, the effectiveness of joint use with state-of-the-art data augmentation is evaluated.

\vspace{-0.15cm}
\subsection{Comparison with test-time augmentation}
\label{sec:exp1}
\vspace{-0.15cm}
{\bf{Datasets.}} We generated 12 different evaluation datasets from the four common image classification datasets (CIFAR10/100, SVHN~\cite{howard2013some}, and CUB-200~\cite{WahCUB_200_2011}) by performing the following three geometrical transformations (Trans.~A, Trans.~B, and Trans.~C):
\vspace{-0.1cm}
\begin{description}
\item[Trans.~A:] 
Randomly rotate the original training and test images to one of four discrete angles (0~[deg], 90~[deg], 180~[deg], and 270~[deg]). This geometrical transformation is a similar dataset to Sec. ~\ref{sec:ana}.
\vspace{-0.1cm}
\item[Trans.~B:] 
Randomly rotate the original training and test images with 
uniformly sampled
angles $\theta \in [0,360]$.
\vspace{-0.1cm}
\item[Trans.~C:] 
Rotate the original training and test images with 
uniformly sampled 
angles $\theta \in [0,360]$, and then scale with random magnifications $r \in [0.5, 1.5]$.
\end{description}
\vspace{-0.1cm}

\noindent
We evaluated the effectiveness of the proposed method using the six comparison methods according to the combination of types of the training data for fine-tuning (original data or geometrically transformed data) and types of TTA (without TTA, the ensemble mean, and the ensemble maximum) as shown in Table~\ref{tbl:Settings}.
For this experiment, we used ResNet18,ResNet50 ~\cite{he2016deep} and MobileNetV2~\cite{sandler2018mobilenetv2}. By default, we train the networks for 200 epochs with mini-batch size 128, weight decay is 0.0005, and momentum 0.9. We used SGD for optimization. The learning rate is set to 0.1 to 0.0001.

\begin{table}[t]
	\begin{center}
		\caption{\bf{Settings  in existing and proposed methods.}}
		\label{tbl:Settings}
		\scalebox{0.825}
		    {
		    \tabcolsep = 0.5mm
			\begin{tabular}{c|c|c|c}
            \multicolumn{2}{c|}{}  & TTA or feature- & {Dataset} \\
            \multicolumn{2}{c|}{}  & map ensemble & for fine-tuning \\
            \hline
            \hline
             & - & without TTA &  \\
            Baseline (BS) & max & Ensemble maximum & original dataset\\
             & mean & Ensemble mean & (e.g., CIFAR10)\\
            \hline
            Data  & - & without TTA &  \\
            Augmentation & max & ensemble maximum & transformed dataset\\
             (DA) & mean & ensemble mean & (e.g., CIFAR10 with Trans.~A)\\
            \hline
            \multicolumn{2}{c|}{Ours (mean)} & ensemble mean & transformed dataset \\
            \multicolumn{2}{c|}{Ours (max) } & ensemble maximum & (e.g., CIFAR10 with Trans.~A) \\
            \end{tabular}
			}
	\end{center}
\end{table}

\noindent
{\bf{Result.}} Table~\ref{tbl:main1} and Table~\ref{tbl:main2} show the results obtained by the proposed and the comparison methods using ResNet and MobilenetV2. 
Bold and underline indicate the highest and next highest values, respectively. 
For all network models and all datasets, the accuracy of the proposed method is higher than those of the comparison methods. 
These results show the effectiveness of the proposed method. Note that the feature map maximum $S_{\rm{max}}$ is generally more effective than the feature map mean $S_{\rm{mean}}$ in the proposed method. Therefore, in the following experiments in Sec.~\ref{sec:exp2} and Sec.~\ref{sec:exp3}, we used the feature maximum $S_{\rm{max}}$ to evaluate the performance of our method.

\begin{table*}[t]
	\begin{center}
		\caption{{\bf{Results for ResNet18 and ResNet50~\cite{he2016deep}}}. Bold and underlines indicate the most and next most accurate methods, respectively.
		ResNet18 was used for CIFAR and SVHN, and ResNet50 was used for CUB-200.}
		\label{tbl:main1}
		\scalebox{0.875}{
    		{\tabcolsep = 1mm
			\begin{tabular}{cc|ccc|ccc|ccc|ccc}
            &  \multicolumn{1}{c|}{} &\multicolumn{3}{c|}{CIFAR10} & \multicolumn{3}{c|}{CIFAR100}  & \multicolumn{3}{c|}{SVHN} & \multicolumn{3}{c}{CUB-200}\\  
            \hline
            & \multicolumn{1}{|c|}{TTA} & Trans.~A & Trans.~B & Trans.~C & Trans.~A & Trans.~B & Trans.~C & Trans.~A & Trans.~B & Trans.~C & Trans.~A & Trans.~B & Trans.~C \\ 
            \hline
            & \multicolumn{1}{|c|}{-} & 55.18	&	48.04	&	40.35	&	43.07	&	31.70	&	24.67	&	43.31	&	42.81	& 36.44 &	52.90	&	50.07	&	46.53	\\
            Baseline& \multicolumn{1}{|c|}{max} &	83.26	&	67.90	&	55.98	&	69.26	&	46.08	&	34.22	&	67.23	&	63.45	& 52.74 &	68.78	&	64.38	&	60.44	\\
            & \multicolumn{1}{|c|}{mean} &	71.53	&	57.30	&	46.67	&	62.81	&	40.88	&	30.47	&	67.47	&	63.24	&  52.6 &	62.70	&	58.23	&	54.68	\\
            \hline
            & \multicolumn{1}{|c|}{-} &	88.82	&	85.15	&	79.29	&	65.05	&	60.95	&	52.91	&	89.66	&	87.41	& 85.17 &	71.28	&	70.87	&	66.12	\\
            fine-tuning& \multicolumn{1}{|c|}{max} &	91.28	&	87.9	&	82.37	&	68.91	&	64.26	&	56.26	&	91.44	&	89.76	& 87.68 &	72.78	&	71.54	&	67.64	\\
            & \multicolumn{1}{|c|}{mean} &	91.55	&	88.28	&	83.05	&	70.82	&	66.37	&	58.31	&	91.73	&	90.39	& 88.43 &	74.65	&	73.84	&	69.05	\\
            \hline
            \multicolumn{2}{c|}{Ours (max)} &	{\bf{94.66}}	&	{\bf{91.86}}	&	{\bf{87.89}}	&	{\bf{75.36}}	&	{\bf{71.13}}	&	{\underline{62.91}}	&	{\underline{92.41}}	&	{\underline{90.83}}	& {\underline{88.70}} &	{\bf{77.51}}	&	{\bf{75.98}}	&	{\bf{72.66}}	\\
            \multicolumn{2}{c|}{Ours (mean)} &	{\underline{94.08}}	&	{\underline{91.76}}	&	{\underline{87.68}}	&	{\underline{75.26}}	&	{\underline{70.45}}	&	{\bf{62.98}}		& {\bf{92.45}} &	{\bf{91.29}} & {\bf{88.80}} &	{\underline{74.23}}	&	{\underline{73.80}}	& {\underline{69.92}}	\\
			\end{tabular}
			}

		}
	\end{center}
	\begin{center}
		\caption{{\bf{Results for MobilenetV2~\cite{sandler2018mobilenetv2}}}. Bold and underlines indicate the most and next most accurate methods, respectively.}
		\label{tbl:main2}
		\scalebox{0.875}{
    		{\tabcolsep = 1mm
			\begin{tabular}{cc|ccc|ccc|ccc|ccc}
            &  \multicolumn{1}{c|}{} &\multicolumn{3}{c|}{CIFAR10} & \multicolumn{3}{c|}{CIFAR100} & \multicolumn{3}{c|}{SVHN} & \multicolumn{3}{c}{CUB-200} \\  
            \hline
            & \multicolumn{1}{|c|}{TTA} & Trans.~A & Trans.~B & Trans.~C & Trans.~A & Trans.~B & Trans.~C & Trans.~A & Trans.~B & Trans.~C & Trans.~A & Trans.~B & Trans.~C \\ 
            \hline
            & \multicolumn{1}{|c|}{-} &	50.47	&	40.61	&	36.40	&	37.44	&	25.05	&	19.81	&	42.21	&	41.67	& 37.95 &	44.44	&	41.63	&	37.07	\\
            Baseline& \multicolumn{1}{|c|}{max} &	82.52	&	59.36	&	50.64	&	61.24	&	36.03	&	28.12	&	78.26	&	69.46	& 62.29 &	55.51	&	50.85	&	45.41	\\
            & \multicolumn{1}{|c|}{mean} &	67.02	&	46.10	&	39.35	&	51.89	&	31.17	&	24.94		&	73.69	&	67.41	& 59.21 &	51.71	&	47.64	&	42.13\\
            \hline
            & \multicolumn{1}{|c|}{-} &	75.75	&	70.10	&	68.89	&	58.66	&	52.93	&	45.73	&	86,00	&	85.27	& 81.82 &	65.02	&	64.67	&	60.68	\\
            fine-tuning& \multicolumn{1}{|c|}{max} &	76.83	&	71.20	&	69.93	&	60.80	&	54.27	&	47.20	&	86.56	&	86.38	& 82.45 &	65.96	&	65.52	&	62.5	\\
            & \multicolumn{1}{|c|}{mean} &	77.94	&	72.17	&	70.7	&	62.14	&	55.36	&	48.58	&	87.73	&	86.91	& 83.44 &	67.31	&	66.95	&	63.65	\\
            \hline
            \multicolumn{2}{c|}{Ours (max)} &	{\underline{82.62}}	&	{\underline{76.28}}	&	{\underline{74.41}}	&	{\bf{65.84}}	&	{\bf{59.48}}	&	{\bf{52.87}}	&	{\bf{89.68}}	&	{\bf{87.72}}	& {\bf{84.65}} &	{\bf{73.58}}	&	{\bf{71.44}}	&	{\bf{67.10}}	\\
            \multicolumn{2}{c|}{Ours (mean)} &	{\bf{85.56}}	&	{\bf{79.60}}	&	{\bf{74.89}}	&	{\underline{64.83}}	&	{\underline{58.59}}	&	{\underline{52.34}}	&	{\underline{89.18}}	&	{\underline{84.45}}	&	{\underline{84.24}} &	{\underline{69.04}}	&	{\underline{67.47}}	&	{\underline{61.67}}	\\
			\end{tabular}
			}
		}
	\end{center}
\end{table*}

\vspace{-0.25cm}
\subsection{Comparison with existing methods using Mnist-rot-12k dataset}
\label{sec:exp2}
\vspace{-0.15cm}
Next, we compared the performance of the existing and proposed methods using MNIST-rot-12k~\cite{larochelle2007empirical}, which is widely used to evaluate the robustness to image rotation. 
This dataset consists of 12k training images and 50k test images. 
In this experiment, ResNet18 and Xavier's initialization was used. 
Other training protocols (e.g., number of mini-batch  size and optimization methods) were the same as those in Sec.~\ref{sec:exp1}.

Table~\ref{tbl:Mnist} shows the performance of the proposed method and the existing methods which employ Maxout architecture. The evaluation results of the state-of-the-art methods are also shown as references. Note that the results of the existing methods are taken from the original paper.
As shown in Table~\ref{tbl:Mnist}, the accuracy of the proposed method is higher than those of the existing method with Maxout architecture~\cite{laptev2016ti,zhang2018mintin}. Furthermore, the performance of the proposed method is comparable or better than the state-of-the-art methods.

\begin{table}[t]
	\begin{center}
	    \vspace{-0.65cm}
		\caption{ {\bf{Result on Mnist-rot-12k.}} Left: Performances of the existing methods using Maxout architecture and the proposed method. Right: Performances of other state-of-the-art methods.}
		\label{tbl:Mnist}
		\scalebox{0.875}{
		    {\tabcolsep = 1.5mm

			\begin{tabular}{cc}
			\begin{tabular}[t]{c|c}
            \multicolumn{1}{c|}{\it{Method}}  & {\it{Acc.}}\\
            \hline
            \hline
            \multicolumn{1}{c|}{TI polling~\cite{laptev2016ti}}	& 97.80 \\
            MINTIN~\cite{zhang2018mintin} & 98.43 \\
            \multicolumn{1}{c|}{{Ours (max)}} & {\bf{98.96}}\\ 
			\end{tabular}
			&
			\begin{tabular}[t]{c|c}
            {\it{Method}}  & {\it{Acc.}}  \\
            \hline
            \hline
            STN~\cite{jaderberg2015spatial}	& 97.12  \\
            GECNN~\cite{cohen2016group}	& 97.72  \\
            Harmonic~\cite{worrall2017harmonic} &	98.40	\\   
            OR-Net~\cite{zhou2017oriented}	& 98.46	\\
            RotEqNet~\cite{marcos2017rotation} & 	98.91 \\
            Xu et.al.~\cite{xu2020towards} & 98.92
			\end{tabular}
			\end{tabular}
			}
		}
    \vspace{-0.3cm}
	\end{center}
\end{table}

\vspace{-0.2cm}
\subsection{Joint use with state-of-the-art data augmentation}
\label{sec:exp3}
Finally, we evaluated the effectiveness of the proposed method in combination with state-of-the-art data augmentation. In this experiment, we used the transformed dataset obtained from the original CIFAR10 and CIFAR100 with Trans.~B described in Sec.~\ref{sec:exp1}.
We used Rand Augmentation (RA)~\cite{cubuk2020randaugment} as state-of-the-art data augmentation.  
The comparison method is also the same as in Sec.~\ref{sec:exp1}. 
Furthermore, we evaluated the performance of the state-of-the-art learning-based TTA (GPS~\cite{molchanov2020greedy}) as a comparison. 
The performance was evaluated using WideResNet~(WR)~\cite{zagoruyko2016wide} as in ~\cite{molchanov2020greedy}. 

The experimental results are shown in Table~\ref{tbl:ra}. 
The proposed method achieves higher accuracy than all existing methods, including GPS~\cite{molchanov2020greedy}. 
These results show that our proposed method can be effectively used in combination with state-of-the-art data augmentation.


\begin{table}[t]
	\begin{center}
	    \vspace{-0.65cm}
		\caption{{\bf{Effectiveness of joint use with state-of-the-art data augmentation.}}}
		\label{tbl:ra}
		\scalebox{0.875}{
			\begin{tabular}{cc|c|c}
            \multicolumn{2}{c|}{} &\multicolumn{1}{c|}{CIFAR10}&\multicolumn{1}{c}{CIFAR100} \\  
            \hline
            \hline
            \multicolumn{2}{l|}{WR~\cite{zagoruyko2016wide}+RA~\cite{cubuk2020randaugment}} & 90.74 & 72.07 \\
            \hline
            \multicolumn{2}{l|}{WR~\cite{zagoruyko2016wide}+RA~\cite{cubuk2020randaugment}+TTA(max)} & 91.68 & 73.51  \\
            \multicolumn{2}{l|}{WR~\cite{zagoruyko2016wide}+RA~\cite{cubuk2020randaugment}+TTA(mean)} & 91.93 & 74.67  \\
            \multicolumn{2}{l|}{WR~\cite{zagoruyko2016wide}+RA~\cite{cubuk2020randaugment}+GPS~\cite{molchanov2020greedy}}  & 92.08 & 74.25  \\
            \multicolumn{2}{l|}{{{WR~\cite{zagoruyko2016wide}+RA~\cite{cubuk2020randaugment}+Ours (max)}}} & {\bf{93.59}} & {\bf{77.50}}
			\end{tabular}
		}
	\end{center}
    \vspace{-0.3cm}
\end{table}

\vspace{-0.2cm}
\section{Conclusions}
\label{sec:intro}
\vspace{-0.2cm}
We have proposed the novel CNN architecture to improve the robustness against geometrical transformations without modifying an existing backbone.
The key is to enclose the existing backbone structure with the geometrical transformation and the feature map ensemble.
We have demonstrated the effectiveness of the proposed method using CIFAR, CUB-200, and Mnist-rot-12k. 
Our future work would be to extend the proposed framework to improve the robustness to photometric transformations.
\end{spacing}

\begin{spacing}{0.915}
\bibliographystyle{IEEEbib}
\bibliography{icip2021_sbt}
\end{spacing}
\end{document}